\documentclass[runningheads]{llncs}
\usepackage[T1]{fontenc}
\usepackage{graphicx}
\usepackage{multirow}
\usepackage{amsmath}

\begin{document}
\title{Federated Learning for Data and Model Heterogeneity in Medical Imaging}
%
%

\author{Hussain Ahmad Madni\inst{1} \and
Rao Muhammad Umer\inst{2} \and
Gian Luca Foresti\inst{1}}
\authorrunning{H. A. Madni et al.}
%
\institute{Department of Mathematics, Computer Science and Physics (DMIF), University of Udine, Udine 33100, Italy.
\and
Institute of AI for Health, Helmholtz Zentrum München - German Research Center for Environmental Health, Neuherberg 85764, Germany.\\
}

\maketitle              
\begin{abstract}
Federated Learning (FL) is an evolving machine learning method in which multiple clients participate in collaborative learning without sharing their data with each other and the central server. In real-world applications such as hospitals and industries, FL counters the challenges of data heterogeneity and model heterogeneity as an inevitable part of the collaborative training. More specifically, different organizations, such as hospitals, have their own private data and customized models for local training. To the best of our knowledge, the existing methods do not effectively address both problems of model heterogeneity and data heterogeneity in FL. In this paper, we exploit the data and model heterogeneity simultaneously, and propose a method, MDH-FL (Exploiting Model and Data Heterogeneity in FL) to solve such problems to enhance the efficiency of the global model in FL. We use knowledge distillation and a symmetric loss to minimize the heterogeneity and its impact on the model performance. Knowledge distillation is used to solve the problem of model heterogeneity, and symmetric loss tackles with the data and label heterogeneity. We evaluate our method on the medical datasets to conform the real-world scenario of hospitals, and compare with the existing methods. The experimental results demonstrate the superiority of the proposed approach over the other existing methods. 
\end{abstract}

\section{Introduction}
Federated Learning (FL), initially introduced by~\cite{mcmahan2017communication}, has become a popular machine learning technique because of distributed model training without sharing the private data of participating hosts. In FL, participants (i.e., clients) including organizations and devices generally have heterogeneous data and heterogeneous models that are customized according to the tasks and local data. In real-world applications, data from multiple sources are heterogeneous and contain non-independent and identically and distributed data (non-IID). Moreover, data from multiple source may produce diverse labels and classes that is more challenging for the convergence of FL model. Traditional training methods based on centralized data cannot be used in practical applications due to privacy concerns and data silos at multiple locations~\cite{kairouz2021advances}. FL has the ability to train a global model by allowing multiple participants to train collaboratively with their decentralized private data. In this way, private data of an individual participant are never shared with the central server and other participant in FL environment. Most common FL algorithms are FedProx~\cite{li2020federated} and FedAvg~\cite{mcmahan2017communication} that aggregate the model parameters obtained from the participating clients. Most of the existing methods~\cite{li2021model,wang2020federated} using these algorithms consider the homogeneous data and same architecture of the local model used by all participants. 
\par
\begin{figure*}[!t]
    \centering
\includegraphics[width=\linewidth]{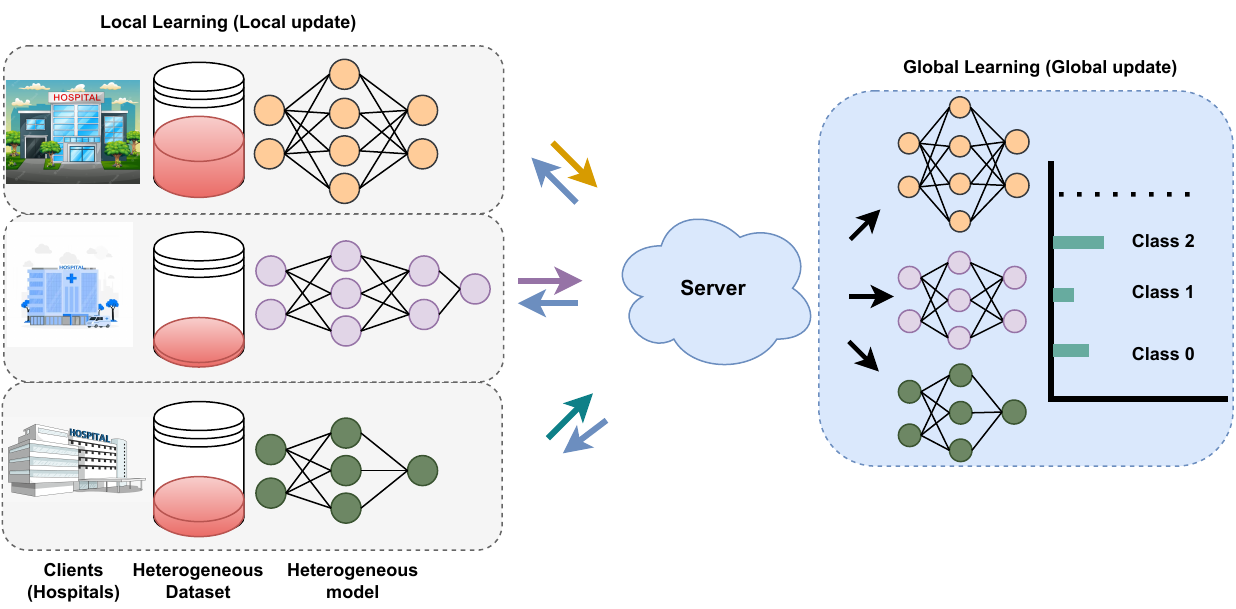}
    \caption{Participating hospitals (i.e., clients) contain heterogeneous local models trained on heterogeneous data and diverse labels. Each client has its own data and custom model as per requirements and tasks.}
    \label{fig:introduction}
\end{figure*}

 
In practical applications, each participant has its own data and might need to design its own customized model~\cite{shen2020federated,jeong2020federated} due to specific and personalized requirements~\cite{li2020federated}. Such heterogeneity in data and model is natural in healthcare organizations that design custom models for specific tasks as illustrated in Fig.~\ref{fig:introduction}. In such environment, hospitals are hesitant to reveal their data and model architecture due to privacy concerns and business matters. Thus, numerous methods have been proposed to perform FL with such heterogeneous data~\cite{huang2021evaluating,warnat2021swarm} and clients~\cite{liang2020think,lin2020ensemble,li2019fedmd}. FedMD~\cite{li2019fedmd} is a method that implemented knowledge distillation based on class scoring calculated by local models of clients trained on public dataset. FedDF~\cite{lin2020ensemble} is another method that performs ensemble distillation by leveraging the unlabeled data for every model architecture. Such existing methods are dependent on shared models and mutual consensus. However, a mutual consensus is another challenge in which each client is unable to set its learning direction to adjust the deviations among all participants. Moreover, designing additional models enhance the processing overhead, and eventually affect the performance. Thus, FL containing heterogeneous data and models without depending on global sharing and consensus is critical and challenging.

\par 
The methods discussed above are mostly dependent on the assumption that every participant has homogeneous, independent and identically distributed (IID) data that is not possible in real-world scenarios. More specifically, in collaborative learning each participant has its own data and requires a customized model for the specific nature of data and task. As FL has many participants with heterogeneous models and data, each model suffers data diversity effecting the overall performance of the global model. Existing methods such as~\cite{ghosh2017robust,van2015learning} have been presented that designed robust loss function to minimize the negative impact of heterogeneous labels in data. The current existing methods either tackle the data heterogeneity or model heterogeneity. In FL, it is required that a model should be robust and learn sufficiently from the data during local update.
\par 
To tackle with the heterogeneous participants and data containing diverse labels are the prominent challenges in FL. Model heterogeneity in FL causes the diverse noise patterns and decision boundaries. Moreover, data heterogeneity based on non-IID and label diversity creates difficulty in convergence of the global model during global learning phase in FL. It is required for each client to concentrate on the contribution of other participants and align the learning to produce a robust global model. In this paper, we propose a solution for the heterogeneous data and model in FL. 1) For the model heterogeneity, a model distribution (i.e., logits output) is aligned by learning the knowledge distribution and feedback from other clients using public data. In this way, each participant learns with its own strategy without depending on the public model. 2) To tackle with the data heterogeneity having diverse labels, an additional symmetric loss function as proposed in~\cite{wang2019symmetric}, is used to minimize the diversity impact on model learning. \\
Our main contribution are as follows.
\begin{itemize}
    \item We explore the real-world scenario of data and model heterogeneity in hospitals implementing decentralized collaborative model training.
    \item We use knowledge distillation for the alignment of model output (i.e., logits) to solve the problem of model heterogeneity and to produce an efficient global model in FL. 
    \item We utilize an additional symmetric loss function to optimize the model learning based-on heterogeneous data containing diverse labels. 
    \item We evaluate the proposed method on hematological cytomorphology clinical datasets with heterogeneous model and data scenarios, and experimental results show the supremacy of the proposed method over the existing FL methods. 
\end{itemize}

\section{Related Work}

\subsection{Federated Learning}
Federated Learning (FL), firstly proposed by~\cite{mcmahan2017communication} is a machine learning method in which multiple clients train a global model without sharing their private local data to preserve the privacy. Initially, FedAvg was used to aggregate the parameters of local models trained on local data~\cite{mcmahan2017communication}. A method similar to FedAvg has been proposed in~\cite{li2020federated} that can customize the local calculations with respect to the iterations and devices used in FL. In~\cite{wang2020federated}, weights of the layers in a client model are collected to accomplish one-layer matching that produce weight of every layer in the global model. Knowledge distillation has been utilized for the communication of FL heterogeneous models in~\cite{li2019fedmd}. In this method, for each client, class scores obtained from the public dataset are collected on the server to calculate the aggregated value to be updated. In~\cite{lin2020ensemble}, unlabeled data leveraging ensemble distillation is used for the model fusion. Global parameters are dynamically assigned as a subset to the local clients according to their capabilities in~\cite{diao2020heterofl}. An algorithm has been introduced in~\cite{liang2020think} to produce a global model from the learning of local representations. We summarize that existing methods assume that all clients have homogeneous data without consideration of any type of heterogeneity. No research have been conducted for the mitigation of diverse impact of data and model heterogeneity simultaneously during the collaborative learning in FL. 

\subsection{Model and Data Heterogeneity}
Numerous methods have been presented to tackle with data heterogeneity, but not much research have been conducted for the model heterogeneity and label diversity in the scenario of FL. Some existing methods use loss functions for the optimization such as~\cite{ghosh2017robust,van2015learning}. A convex classification calibration loss has been proposed by~\cite{van2015learning} that is robust for incorrect classes and labels. Some loss functions are evaluated by~\cite{ghosh2017robust} that prove the robustness of MAE to perturbed classes in deep learning. Estimation of the probability for every class flipped to some other class has been utilized in existing methods~\cite{yao2019safeguarded,patrini2017making,sukhbaatar2014training}. In~\cite{yao2019safeguarded}, corrupted data is transformed into Dirichlet distribution space and label regression technique is used to infer the correct classes, and finally data modeling and classifier are trained together. Some existing methods extract clean samples, re-weight each instance, or apply some transformation on the heterogeneous data for model training~\cite{jiang2018mentornet,wei2020combating,han2018co}.
\par 
 A method JoCoR has been proposed in~\cite{wei2020combating} that uses Co-Regularization for the joint loss estimation. In this method, the samples with minimum loss are selected to update the model parameters. MentorNet is another method proposed by~\cite{jiang2018mentornet} that comes up with a technique used to weight a sample such as used in StudentNet and MentorNet. Co-teaching method has been proposed in~\cite{han2018co} that selects data for cross training of the two deep networks simultaneously. To avoid the model overfitting specific samples, robust regularization is used in~\cite{miyato2018virtual,arpit2017closer,zhang2017mixup}. A method Mixup has been proposed in~\cite{zhang2017mixup} to regularize the deep network by training the convex pairs of instance and their corresponding labels. Regularization is used by~\cite{arpit2017closer} to minimize the impact of corrupted data while not affecting the training of actual samples. A regularization method has been introduced in~\cite{miyato2018virtual} that depends on the virtual adversarial loss and adversarial direction that do not require any label information. Most of the existing methods that solve he problem of data heterogeneity and corrupted data are based on centralized data and a single model. However, server is not able to access the local data of a client directly in FL environment. Moreover, heterogeneous clients have diverse patterns and decision boundaries. 

\section{Federated Learning with Heterogeneous Data and Models}
\begin{figure*}[!t]
    \centering
    \includegraphics[width=\linewidth]{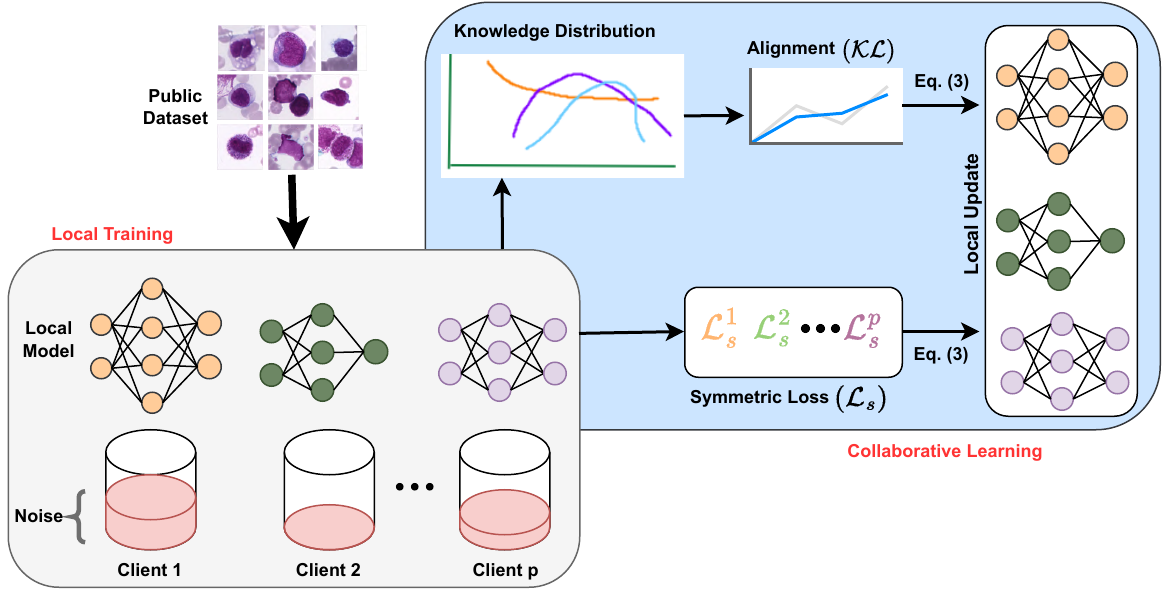}
    \caption{Proposed approach containing local training and global learning in FL. Local models are updated with Kullback-Leibler loss based on knowledge distribution (Eq.~\ref{eq:equation3}), and Symmetric loss (Eq.~\ref{eq:equation8}). In local training phase, private models are individually trained on private datasets, and in global or collaborative learning, local clients are updated through loss functions (i.e., $\mathcal{KL}$ and $\mathcal{L_s}$).} 
    \label{fig:methodology}
\end{figure*}

In FL with heterogeneous participants $P$ and a server, we consider $C$ as the number of all clients where $|C| = P$. Thus, the $p^{th}$ participant $c_{p} \in C$ has its local data $d_{p} = \{(x_{i}^{p}, y_{i}^p) \}_{i = 1}^{N_{p}}$ where $|x|^{p} = N_{p}$. Moreover, $y_{i}^{p} \in \{0, 1 \}^{N_{p}}$ is a one hot vector containing ground truth labels. Furthermore, a local model $\Theta_{p}$ owned by a client $c_{p}$ has different architecture and $f(x^{p}, \Theta_{p})$ represents the logits output produced by the network $f(.)$ using input $x^{p}$ calculated with $\Theta_{p}$. The server has a public dataset $d_{0} = \{x_{i}^{0}\}_{i = 1}^{N_{0}}$ that may belongs to the client data for different classification tasks. In FL, overall process is divided into local training and collaborative learning in which local training is performed by $E_{l}$ rounds and collaborative learning is performed by $E_{c}$ rounds. Our purpose is to perform FL with heterogeneous (i.e., non-IID) data containing diverse labels and heterogeneous clients, so a client has its heterogeneous data $\Tilde{d} = \{(x_{i}^{p}, \Tilde{y}_{i}^{p})\}_{i = 1}^{N_{p}}$ in which $\Tilde{y}_{i}^{p}$ denotes the heterogeneous annotations. Each client has different noise patterns and decision boundaries due to model heterogeneity that can be expressed as $f(x, \Theta_{p_1}) \neq f(x, \Theta_{p_2})$. Thus, each client $c_{p}$ must also consider the heterogeneity of other clients $c_{p_0} \neq p$, other than heterogeneity in its own dataset. The overall objective is to find an optimal solution for model parameters $\Theta_{p} = argmin~\mathcal{L}(f(x^{p}, \Theta_{p}), y^{p})$. The architecture of the proposed method is shown in Fig.~\ref{fig:methodology}. Each client is trained on its private dataset and subsequently on public dataset to use the knowledge distillation and alignment of knowledge distribution as given in Eq. (\ref{eq:equation3}). Moreover, each local client is updated and optimized using symmetric loss given in Eq. (\ref{eq:equation8}). 

\subsection{Model Heterogeneity}
The knowledge distribution represented as $D_{p}^{t_c} = f(d_{0}, \Theta_{p}^{t_c})$ is produced for the client $c_{p}$. To estimate the variance in knowledge distribution, Kullback-Leibler ($\mathcal{KL}$) divergence is used by each client as proposed in~\cite{fang2022robust}. $KL$ divergence represents the deviation between two probability distributions. If there are two clients $c_{p_1}$ and $c_{p_2}$ having knowledge distributions $D_{p_1}^{t_c} = f(d_{0}, \Theta_{p1}^{t_c})$ and $D_{p_2}^{tc} = f(d_{0}, \Theta_{p_2}^{t_c})$ respectively, then the difference between their knowledge distributions can be formulated as:
\begin{equation}
\label{eq:equation1}
    \mathcal{KL}(D_{p_1}^{e_c} || D_{p_2}^{e_c}) = \sum D_{p_1}^{e_c} \log (\frac{D_{p_1}^{e_c}}{D_{p_2}^{e_c}})
\end{equation}

If the difference between two knowledge distributions $D_{p_1}^{t_c}$ and $D_{p_2}^{t_c}$ is higher, there is more opportunity for the clients $c_{p_1}$ and $c_{p_2}$ to learn from each other, and vice versa. If the $\mathcal{KL}$ difference is minimized due to the probability distributions $D_{p_1}^{t_c}$ and $D_{p_2}^{t_c}$, it is assumed a technique that allows a client $c_{p_1}$ for learning from the client $c_{p_2}$. Thus, knowledge distribution difference for a client $c_{p}$ can be expressed as:
\begin{equation}
\label{eq:equation2}
    \mathcal{L}_{pl}^{p, e_c} = \sum_{p_0 = 1, p_0 \neq p}^{P} \mathcal{KL}(D_{p_0}^{e_c} || D_{p}^{e_c})
\end{equation}
where $p_{0}$ is a participant other than $c_{p}$. Moreover, knowledge distribution difference is calculated for a client $c_{p}$, so other participants can access the knowledge of $c_{p}$ without leakage of model architecture and data privacy. All participants are prompted for the collaborative learning due to significant difference in their knowledge distributions. Thus, each participant aligns its knowledge distribution by learning from other participants. This process can be mathematically formulated as follows. 
\begin{equation}
\label{eq:equation3}
    \Theta_{p}^{e_c} \leftarrow \Theta_{p}^{e_{c} - 1} - \alpha \nabla_{\Theta}(\frac{1}{P - 1} \cdot \mathcal{L}_{pl}^{p, e_{c} - 1})
\end{equation}
Where $\alpha$ represents the learning rate.

\subsection{Data and Labels Heterogeneity}
We utilize the Symmetric Cross Entropy proposed in~\cite{wang2019symmetric} to minimize the effect of local noise in model learning. Cross Etnropy (CE) is a very common loss function used in most of the classification tasks. CE is deformation of $\mathcal{KL}$ divergence, so $\mathcal{KL}$ can be formulated in term of CE. For example, if $p$ and $g$ are predicted and label class distribution respectively, the $\mathcal{KL}$ divergence can be formulated as:
\begin{equation}
    \mathcal{KL}(g || p) = \underbrace{\sum g(x) \log (g(x))}_{\text{entropy of g}} - \underbrace{\sum g(x) log (p(x))}_{\text{cross entropy}}
    \label{eq:kl_ce}
\end{equation}
The equation~\eqref{eq:kl_ce} contains entropy of $g$ and cross entropy terms. Thus, CE loss for the input $x$ is represented as:
\begin{equation}
\label{eq:equation5}
    \mathcal{L}_{c} = - \sum_{i = 1}^{N} g(x_i) \log (p(x_i))
\end{equation}
Cross Entropy loss ($\mathcal{L}_{c}$) has limitations due to label noise. It does not make overall classes to learn enough from all categories due to various simplicity levels in the classes. To converge the model for such difficult classes, extra communication rounds are required for additional learning. In such scenario, there is a possibility of overfitting to the heterogeneous labels that eventually reduces the overall efficiency of the model. 
\par 
Generally, a model has limited ability for some categories to classify correctly. Moreover, a model prediction is reliable up to some extent due label noise. Thus, if $g$ is not a real class distribution, reliability of prediction $p$ as a true class distribution is limited. To solve this problem, a Reverse Cross Entropy (RCE) loss function proposed in~\cite{wang2019symmetric}, on the basis of $p$ is exploited to align the predicted class distribution by the model. RCE loss for the input $x$ is formulated as:
\begin{equation}
\label{eq:equation6}
    \mathcal{L}_{rc} = - \sum_{i = 1}^{N} p(x_i) \log (g(x_i))
\end{equation}
\par 
It is feasible to learn the difficult classes for the model if both $\mathcal{L}_c$ and $\mathcal{L}_{rc}$ are combined, and overfitting can be avoided. This combined loss is named as Symmetric loss~\cite{wang2019symmetric} that can be expressed as:
\begin{equation}
\label{eq:equation7}
    \mathcal{L}_{s} = \lambda \mathcal{L}_{c} + \mathcal{L}_{rc}
\end{equation}
Where $\lambda$ is used to control the overfitting to noise. Thus, $\mathcal{L}_{c}$ fits the model on each class and $\mathcal{L}_{rc}$ tackles with the label noise. 
\par 
A client aligns its local knowledge with the knowledge of other participants using a local learning process. A local model updated with its local data to prevent the local knowledge forgetting. In the process of local training, label noise redirects the model to wrong learning that causes convergence failure for the model. To solve this problem, symmetric loss ($\mathcal{L}_{s}$) is used to compute the loss between given label and the predicted pseudo-label by the model. Local update for a model can be expressed as:
\begin{equation}
\label{eq:equation8}
    \Theta_{p}^{el} \leftarrow \Theta_{p}^{e_{l} - 1} - \alpha \nabla_{\Theta} \mathcal{L}_{s}^{p, e_{l} - 1} (f(x^{p}, \Theta_{p}^{e_{l} - 1}), \Tilde{y}^{p})
\end{equation}
where $e_{p} \in E_{l}$ denotes the $e_{p}$-th epoch in model training. A client leverages $\mathcal{L}_{s}$ to update its model that strengthens the local knowledge, and avoids the overfitting to label noise. Thus, model learning is promoted with the $\mathcal{L}_{s}$ loss.

\begin{table}[t]
    \centering
    \begin{tabular}{c|c}
        \bfseries Hyperparameter & \bfseries Value(s)  \\
        \hline
        $E_c$ (global epochs for collaborative learning) & 40 \\ 
        $E_l$ (local epochs for local training) & $\frac{N_p}{N_0}$ \\
        Optimizer & Adam~\cite{kingma2014adam} \\
        $\alpha$ (Initial learning rate for optimizer) & $0.001$ \\
        $b$ (batch size) & $16$ \\
        $\lambda$ & $0.1$ \\
        $\mu$ (labels diversity rate) & $\{0.1, 0.2, 0.3\}$ \\
        flip \%age in data $\Tilde{d}$ & $20$ \\
        $\gamma$ (data heterogeneity rate) & $0.5$ \\
        \hline
    \end{tabular}
    \caption{Federated Learning hyperparameters.}
    \label{tab:configuration}
\end{table} 

\section{Experimental Results}
\subsection{Datasets and Models}
In the experiments, two hematological cytomorphology clinical datasets, INT\_20 dataset~\cite{umer2023imbalanced} and Matek\_19 dataset~\cite{matek2019human} are used for the single-cell classification in Leukemia (i.e., cancer detection). INT\_20 dataset~\cite{umer2023imbalanced} is used as a public dataset on the server, and Matek\_19 dataset~\cite{matek2019human} is distributed to the clients as their local private datasets. INT\_20 dataset has $26379$ samples of $13$ classes containing $288 \times 288$ colored blood images. Matek\_19 dataset contains a total of $14681$ samples of 13 classes having blood images with resolution of $400 \times 400$. In each experiment, four clients are set up for the collaborative learning and Matek\_19 dataset is equally divided to these clients using Dirichlet distribution (i.e., Dir ($\gamma$)) to make non-IID dataset~\cite{madni2023blockchain}. The size of public data on the server and private data on each client is $N_0 = 26379$ and $N_p = 3670$ respectively. 
\par 
For the homogeneous clients, ResNet-12~\cite{he2016deep} is used for the training of all clients, and for the heterogeneous scenario, ShuffleNet~\cite{zhang2018shufflenet}, ResNet10~\cite{he2016deep}, Mobilenetv2~\cite{sandler2018mobilenetv2}, and ResNet12~\cite{he2016deep} are assigned to the clients for local training on the private datasets. 
\par 
To produce labels diversity in data, a matrix $\mathcal{M}$ for the label transition is used represented as $M_{ij} = flip(\Tilde{y} = j|y = i)$ that shows that label $y$ is moved to a heterogeneous class $j$ from a class $i$ . As the real-word scenario, a client $c_p$ selects $N_p$ examples randomly from the private data (Matek\_19), so each client has different noise proportion in its local data. Pair flip~\cite{han2018co} and Symmetric flip~\cite{van2015learning} are the two common categories of matrix $\mathcal{M}$. In Pair flip, a label of original class is swapped with a same wrong category, and in Symmetric flip, a class label is swapped with a wrong class label having same probability. Other implementation configuration is given in Table~\ref{tab:configuration}. 
\par 
In Table~\ref{tab:configuration}, $E_c$ is used as epoch for global or collaborative learning, and $E_l$ is used as local epoch in local training. Adam~\cite{kingma2014adam} is used as optimizer with the learning rate $\alpha$. In each experiment, $\lambda = 0.1$ is fixed to control the overfitting to label diversity. Different diversity rate $\mu$ is used to check the performance of the model with varying data and label heterogeneity. Moreover, label flip percentage for the heterogeneous data $\Tilde{d}$ is fixed as $20$ in all the experiments, where $\gamma = 0.5$ is the data heterogeneity rate. 


\subsection{Comparison with state-of-the-art methods}
We perform experiments to evaluate and compare the proposed method with existing methods on the basis of accuracy. Table~\ref{tab:no_noise} shows the results of different methods using non-heterogeneous training models with $\mu = 0$ (i.e., no label diversity) in local datasets. Performance of each individual client is given in terms of accuracy (\%age), and in the last column average accuracy is given for each method. It is evident that the proposed method performs better when using non-heterogeneous models and homogeneous data without label diversity for model training. 
\par 
Table~\ref{tab:noise} shows the comparison of the proposed method with similar existing methods. We use different labels-diversity techniques for the datasets used with heterogeneous models for the training. Performance of each method is decreased with the increasing labels-diversity rate. Moreover, there is a remarkable difference among all methods when the type of labels diversity is changed. This is because heterogeneous data or labels lead to wrong learning and communication of participating clients. Moreover, heterogeneous models produce different noise patterns that eventually decrease the model performance. Results from the Table~\ref{tab:no_noise} and Table~\ref{tab:noise} are computed when using heterogeneous models. However, these results are computed from the experiments without using additional loss functions. 

\begin{table}[t]
    \centering
    \begin{tabular}{c|c|c|c|c}
         \bfseries Method & \bfseries $\Theta_1$ & \bfseries $\Theta_2$ & \bfseries $\Theta_3$ & \bfseries Average \\
         \hline
         SL-FedL~\cite{madni2023blockchain} & $74.34$ & $77.45$ & $79.23$ & $77.01$ \\
         FedDF~\cite{lin2020ensemble} & $75.54$ & $79.45$ & $77.96$ & $77.65$ \\
         Swarm-FHE~\cite{madni2023swarm} & $76.58$ & $72.98$ & $73.89$ & $74.48$ \\
         FedMD~\cite{li2019fedmd} & $77.83$ & $78.11$ & $78.05$ & $78.00$ \\
         Ours & $\mathbf{82.27}$ & $\mathbf{79.24}$ & $\mathbf{83.57}$ & $\mathbf{81.69}$  \\
         \hline
    \end{tabular}
    \caption{Results produced by FL training with heterogeneous models trained on homogeneous data containing homogeneous labels.}
    \label{tab:no_noise}
\end{table}

\begin{table}[t]
    \centering
    \resizebox{1.0\textwidth}{!}{
    \begin{tabular}{c|c|c|c|c|c|c|c|c|c}
        & & \multicolumn{4}{c|}{\bfseries Symmetric flip} & \multicolumn{4}{c}{\bfseries Pair flip} \\
         \cline{3-10}
         \bfseries Noise Rate ($\mu$) & \bfseries Method & \bfseries $\Theta_1$ & \bfseries $\Theta_2$ & \bfseries $\Theta_3$ & \bfseries Average & \bfseries $\Theta_1$ & \bfseries $\Theta_2$ & \bfseries $\Theta_3$ & \bfseries Average \\
         \hline
         \multirow{4}{*}{$0.1$} & SL-FedL~\cite{madni2023blockchain} & $69.48$ & $71.89$ & $74.59$ & $71.99$ & $70.04$ & $71.98$ & $75.55$ & $72.52$ \\
         & FedDF~\cite{lin2020ensemble} & $71.77$ & $74.45$ & $72.47$ & $72.90$ & $73.02$ & $74.58$ & $73.24$ & $73.61$ \\
         & Swarm-FHE~\cite{madni2023swarm} & $71.15$ & $67.08$ & $67.54$ & $68.59$ & $71.89$ & $67.66$ & $71.93$ & $70.49$ \\
         & FedMD~\cite{li2019fedmd} & $72.88$ & $75.76$ & $73.37$ & $74.00$ & $73.17$ & $75.91$ & $74.42$ & $74.50$ \\
         & Ours & $\mathbf{79.38}$ & $\mathbf{76.95}$ & $\mathbf{79.36}$ & $\mathbf{78.56}$ & $\mathbf{80.04}$ & $\mathbf{77.22}$ & $\mathbf{79.78}$ & $\mathbf{79.01}$ \\
         \hline
        \multirow{4}{*}{$0.2$} & SL-FedL~\cite{madni2023blockchain} & $66.53$ & $68.23$ & $71.15$ & $68.64$ & $66.27$ & $68.76$ & $71.88$ & $68.97$\\
         & FedDF~\cite{lin2020ensemble} & $68.65$ & $70.14$ & $68.64$ & $69.14$ & $69.94$ & $70.09$ & $69.60$ & $69.88$ \\
         & Swarm-FHE~\cite{madni2023swarm} & $65.35$ & $62.89$ & $62.45$ & $63.56$ & $65.82$ & $63.56$ & $68.26$ & $65.88$ \\
         & FedMD~\cite{li2019fedmd} & $67.22$ & $70.32$ & $69.10$ & $68.88$ & $69.18$ & $70.65$ & $71.84$ & $70.56$ \\
         & Ours & $\mathbf{74.06}$ & $\mathbf{71.77}$ & $\mathbf{73.94}$ & $\mathbf{73.26}$ & $\mathbf{78.27}$ & $\mathbf{75.44}$ & $\mathbf{76.67}$ & $\mathbf{76.79}$ \\
         \hline
         \multirow{4}{*}{$0.3$} & SL-FedL~\cite{madni2023blockchain} & $62.14$ & $65.06$ & $66.85$ & $64.68$ & $62.16$ & $64.78$ & $66.76$ & $64.57$ \\
         & FedDF~\cite{lin2020ensemble} & $62.87$ & $66.23$ & $62.97$ & $64.02$ & $66.09$ & $67.35$ & $67.58$ & $67.01$ \\
         & Swarm-FHE~\cite{madni2023swarm} & $59.44$ & $55.91$ & $54.34$ & $56.56$ & $59.96$ & $58.75$ & $61.16$ & $59.96$ \\
         & FedMD~\cite{li2019fedmd} & $61.87$ & $63.78$ & $64.93$ & $63.53$ & $67.45$ & $66.33$ & $67.48$ & $67.09$ \\
         & Ours & $\mathbf{66.54}$ & $\mathbf{65.60}$ & $\mathbf{66.20}$ & $\mathbf{66.11}$ & $\mathbf{73.25}$ & $\mathbf{68.12}$ & $\mathbf{69.12}$ & $\mathbf{70.16}$ \\
         \hline
    \end{tabular}}
    \caption{FL training results computed on heterogeneous models and heterogeneous data for different methods.}
    \label{tab:noise}
\end{table}

We compare the proposed method with similar baseline methods, SL-FedL~\cite{madni2023blockchain}, FedDF~\cite{lin2020ensemble}, Swarm-FHE~\cite{madni2023swarm}, and FedMD~\cite{li2019fedmd}. In the experiments, different diversity rate (i.e., $\mu = \{0.1, 0.2, 0.3\}$) and types are used for the fair comparison. Symmetric loss is used to optimize the model against data and label diversity. Moreover, knowledge distribution (i.e., $\mathcal{KL}$ loss) is implemented to align the output of all heterogeneous participants. Table~\ref{tab:comparison} shows the comparison of proposed method with the existing methods, and demonstrates the outperforming of the proposed method compared to the existing similar methods. 

\begin{table}[t]
    \centering
    \begin{tabular}{c|c|c|c|c|c|c}
         \bfseries Method & \multicolumn{3}{c|}{\bfseries Symmetric flip} & \multicolumn{3}{c}{\bfseries Pair flip}  \\
         \cline{2-7} 
         & \bfseries $\mu = 0.1$ & \bfseries $\mu = 0.2$ & \bfseries $\mu = 0.3$ & \bfseries $\mu = 0.1$ & \bfseries $\mu = 0.2$ & \bfseries $\mu = 0.3$  \\
         \hline
         SL-FedL~\cite{madni2023blockchain} & $76.21$ & $72.16$ & $67.02$ & $78.24$ & $73.87$ & $68.44$ \\
         FedDF~\cite{lin2020ensemble} & $78.53$ & $74.47$ & $68.77$ & $78.91$ & $74.22$ & $69.65$ \\
         Swarm-FHE~\cite{madni2023swarm} & $72.44$ & $66.88$ & $59.94$ & $73.68$ & $67.20$ & $60.72$ \\
         FedMD~\cite{li2019fedmd} & $79.78$ & $74.18$ & $68.11$ & $80.85$ & $76.15$ & $73.26$\\
         Ours & $\mathbf{83.69}$ & $\mathbf{79.82}$ & $\mathbf{72.93}$ & $\mathbf{84.06}$ & $\mathbf{80.10}$ & $\mathbf{73.94}$ \\
         \hline
    \end{tabular}
    \caption{Training results computed with different methods. Heterogeneous models and data are used for each experiment. Two losses (i.e., $\mathcal{L}_{s}$ and $\mathcal{KL}$) are used to minimize the heterogeneity impact and to improve the overall performance of the global model.}
    \label{tab:comparison}
\end{table}

\section{Conclusion}
In this paper, a real-world problem of model and data heterogeneity in medical imaging has been explored. To solve the problem of heterogeneous data and labels diversity, an additional symmetric loss has been used to optimize the model trained on local and private data. To tackle with the heterogeneous participants in FL, Kullback-Leibler has been exploited to align the different noise patterns produced by the heterogeneous participants. Moreover, each participating client uses the knowledge distribution of other participants to improve the performance of global FL model. Experimental results conclude that the proposed method outperforms the existing similar methods. 

\subsubsection{Acknowledgements} This work was supported by the Departmental Strategic Plan (PSD) of the University of Udine Interdepartmental Project on Artificial Intelligence (2020–2025).

%
%
\newpage
\bibliographystyle{splncs04}
\bibliography{references}

\end{document}